\documentclass{article}

\usepackage[nonatbib,preprint]{neurips_data_2024}

\usepackage{fancyhdr}
\usepackage[numbers]{natbib}
\setcitestyle{numbers}

\usepackage[utf8]{inputenc} 
\usepackage[T1]{fontenc}    
\usepackage{url}            
\usepackage{booktabs}       
\usepackage{amsfonts}       
\usepackage{nicefrac}       
\usepackage{microtype}      
\usepackage{amsmath}
\usepackage{amssymb}
\usepackage{mathtools}
\usepackage{amsthm}
\usepackage{multirow}
\usepackage{hyperref}
\usepackage[table,dvipsnames]{xcolor}
\usepackage{babel}
\usepackage{subcaption}
\usepackage{makecell}
\usepackage{xspace}
\usepackage{siunitx}
\usepackage{caption} 

\hypersetup{
  colorlinks   = true, 
  urlcolor     = RoyalBlue, 
  linkcolor    = RoyalBlue, 
  citecolor   = RoyalBlue 
}
\usepackage{pifont}
\definecolor{deepgreen}{RGB}{0, 100, 0}
\newcommand{\cmark}{\color{deepgreen}\checkmark}
\newcommand{\xmark}{\color{red}\ding{53}}

\definecolor{light-light-gray}{gray}{0.92} 
\usepackage{rotating}
\usepackage{placeins}
\usepackage{enumitem}
\usepackage{wrapfig}

\usepackage{colortbl}
\usepackage{cellspace}
\usepackage{boldline, hhline}

\newcommand{\ours}{FlashRAG}
\newcommand{\huggingface}{\raisebox{-1.5pt}{\includegraphics[height=1.05em]{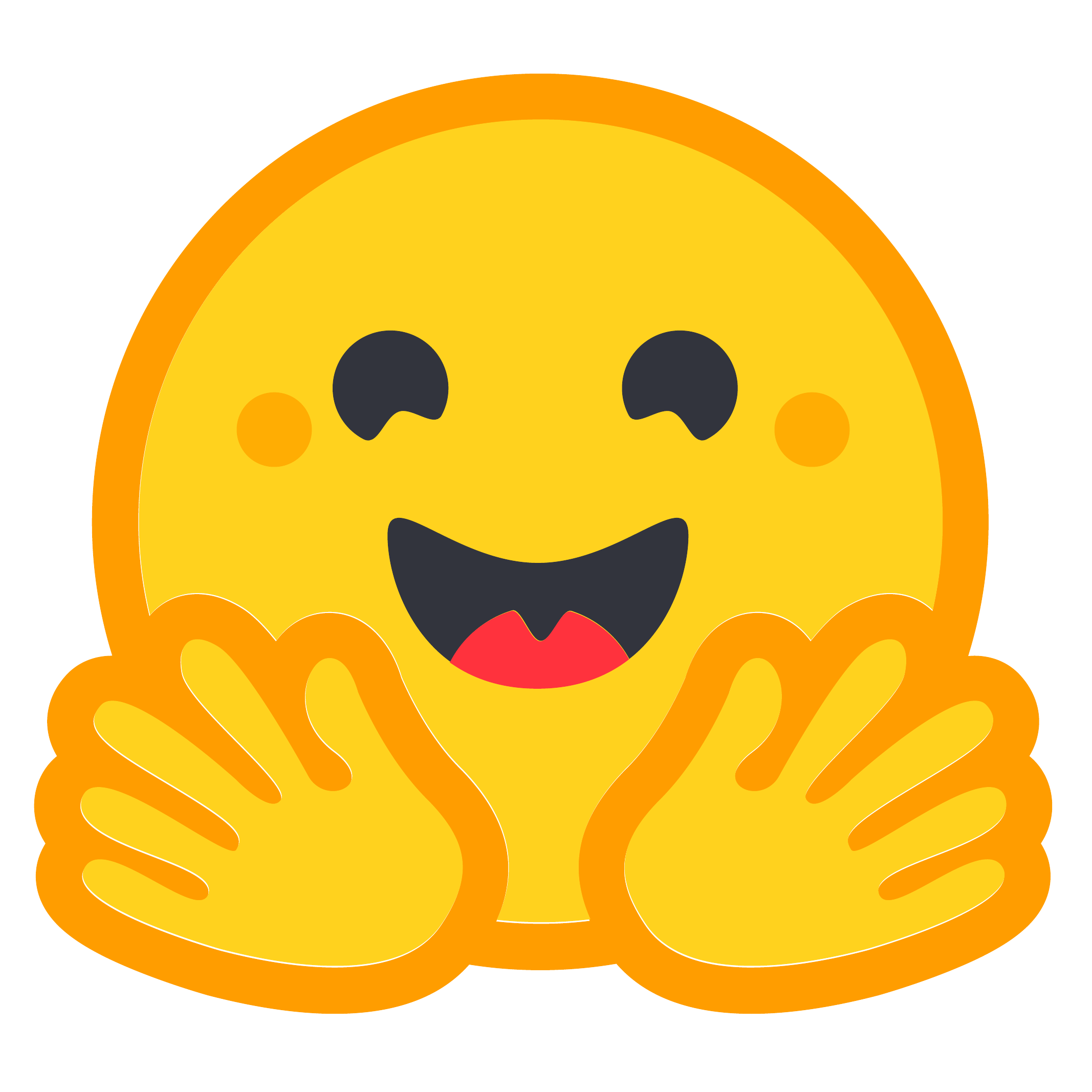}}\xspace}
\newcommand{\github}{\raisebox{-1.5pt}{\includegraphics[height=1.05em]{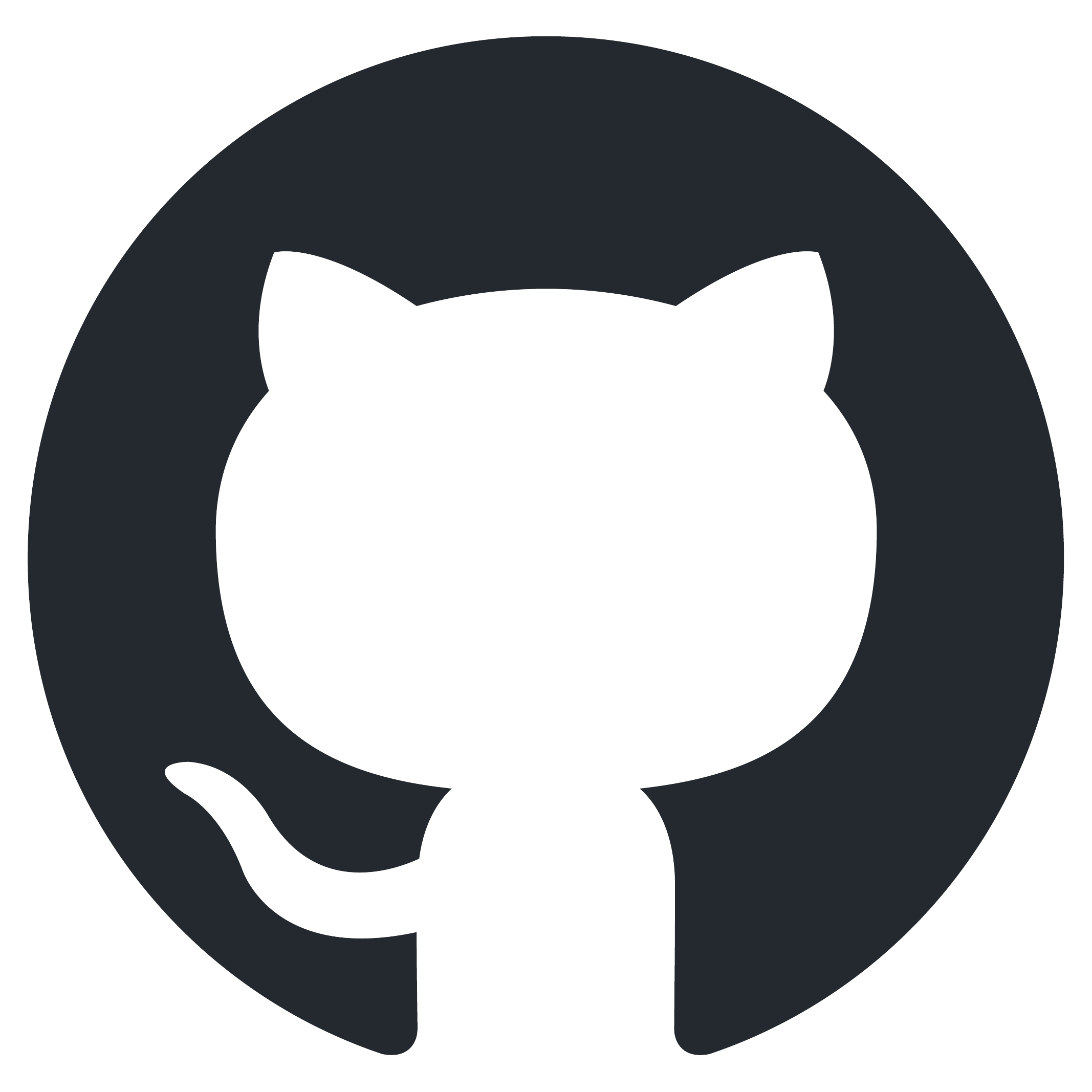}}\xspace}
\newcommand{\modelscope}{\raisebox{0pt}{\includegraphics[height=0.8em]{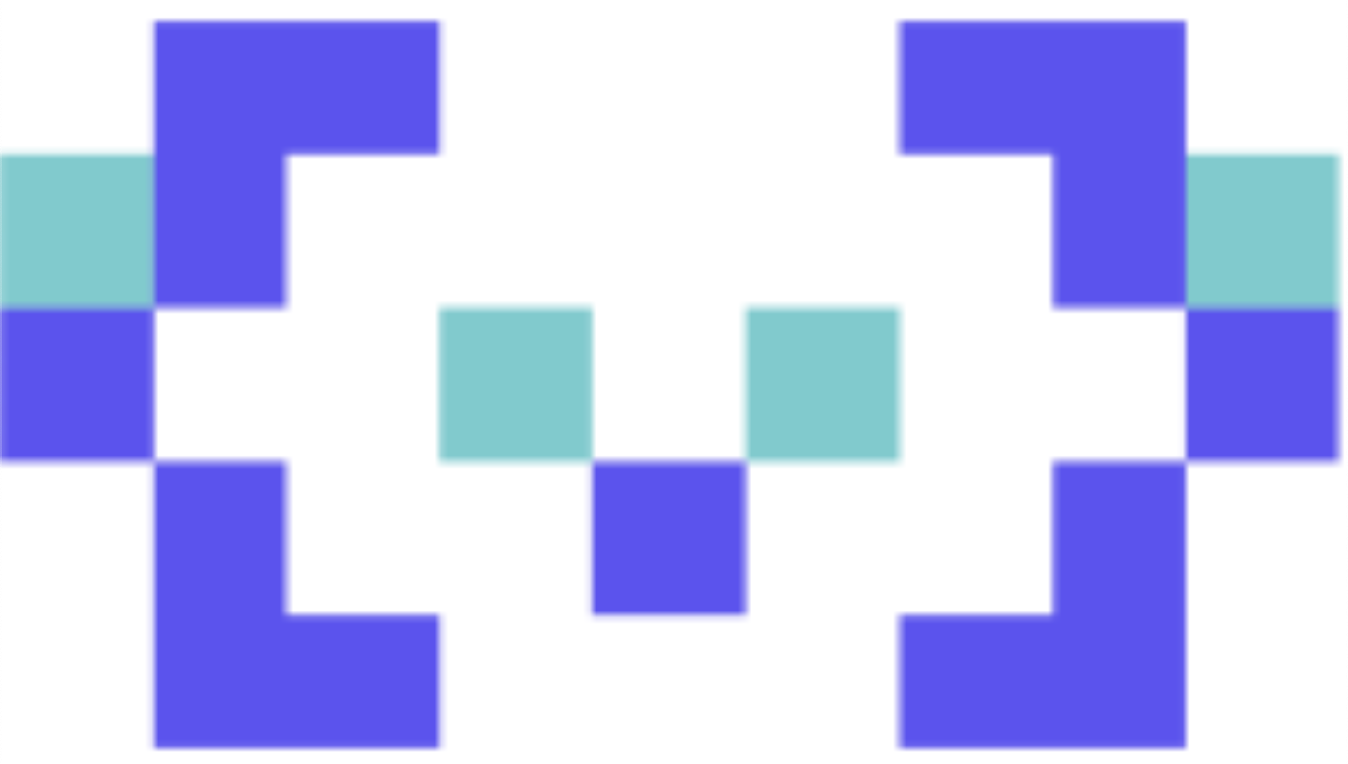}}\xspace}
\newcommand{\project}{\raisebox{0pt}{\includegraphics[height=1.0em]{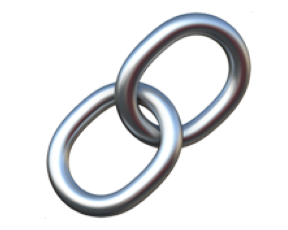}}\xspace}

\newcolumntype{w}{>{\columncolor{white}}c}
\usepackage{caption}

\usepackage{marginnote}

\usepackage[capitalize,noabbrev]{cleveref}

\usepackage{xpatch}
\makeatletter
\renewcommand\paragraph{\@startsection{paragraph}{4}{\z@}                                     {1.35ex \@plus1ex \@minus.2ex}                                {-.5em}
{\normalfont\normalsize\bfseries}}
\makeatother

\title{\raisebox{-0.08cm}{\includegraphics[height=0.6cm]{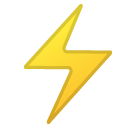}}\ours{}: A Modular Toolkit for Efficient Retrieval-Augmented Generation Research}

\author{%
  Jiajie Jin \quad Yutao Zhu$^*$ \quad Guanting Dong \quad Yuyao Zhang \quad Xinyu Yang  \\ \textbf{Chenghao Zhang \quad Tong Zhao \quad Zhao Yang \quad Zhicheng Dou\thanks{Corresponding authors.} \quad Ji-Rong Wen} \\
  Gaoling School of Artificial Intelligence, Renmin University of China\\
  \texttt{\{jinjiajie, dou\}@ruc.edu.cn, yutaozhu94@gmail.com} \\
  \vspace{.5em}\project\href{https://ruc-nlpir.github.io/FlashRAG/}{Homepage} 
    \vspace{.5em}\huggingface \href{https://huggingface.co/datasets/RUC-NLPIR/FlashRAG_datasets}{Datasets} 
    \vspace{.5em}\modelscope \href{https://www.modelscope.cn/datasets/hhjinjiajie/FlashRAG_Dataset}{Datasets} 
    \vspace{.5em}\github \href{https://github.com/RUC-NLPIR/FlashRAG}{Code}
}

\begin{document}

\maketitle

\begin{abstract}

With the advent of large language models (LLMs) and multimodal large language models (MLLMs), the potential of retrieval-augmented generation (RAG) has attracted considerable research attention. Various novel algorithms and models have been introduced to enhance different aspects of RAG systems. However, the absence of a standardized framework for implementation, coupled with the inherently complex RAG process, makes it challenging and time-consuming for researchers to compare and evaluate these approaches in a consistent environment. Existing RAG toolkits, such as LangChain and LlamaIndex, while available, are often heavy and inflexibly, failing to meet the customization needs of researchers. In response to this challenge, we develop \ours{}, an efficient and modular open-source toolkit designed to assist researchers in reproducing and comparing existing RAG methods and developing their own algorithms within a unified framework. Our toolkit has implemented 16 advanced RAG methods and gathered and organized 38 benchmark datasets. It has various features, including a customizable modular framework, multimodal RAG capabilities, a rich collection of pre-implemented RAG works, comprehensive datasets, efficient auxiliary pre-processing scripts, and extensive and standard evaluation metrics.


\end{abstract}

\section{Introduction}
In the era of large language models (LLMs), retrieval-augmented generation (RAG)~\cite{borgeaud2022retro,guu2020realm} has emerged as an effective solution to mitigate hallucination issues by leveraging external knowledge bases~\cite{bang2023hallucination,chatkbqa,skp}. The substantial applications and potential of RAG have attracted considerable research attention~\cite{rag_survey_gao, coral_yiruo,dparag}. However, with the introduction of a large number of new algorithms and models in recent years, comparing and evaluating these methods in a consistent setting has become increasingly challenging.

Many existing methods are not open-source or require specific configurations for implementation, making adaptation to custom data and innovative components challenging. The datasets and retrieval corpora often vary, with resources being scattered and requiring considerable pre-processing efforts. Besides, the inherent complexity of RAG systems, which involve indexing, retrieval, and generation, often demands extensive technical implementation. While there are some existing RAG toolkits such as LangChain~\cite{LangChain} and LlamaIndex~\cite{LlamaIndex_2022}, they are typically complex and cumbersome, restricting researchers from tailoring processes to their specific needs. Therefore, there is a clear demand for a unified, research-focused RAG toolkit to simplify method development and facilitate comparative studies.

To address the aforementioned issues, we introduce \ours{}, an open-source library that empowers researchers to reproduce, benchmark, and innovate within the RAG domain efficiently. This library offers built-in pipelines for replicating existing work, customizable components for crafting tailored RAG workflows, and streamlined access to organized datasets and corpora to accelerate research processes. \ours{} provides a more researcher-friendly solution compared to existing toolkits. To summarize, the key features of our \ours{} library include:

\textbf{A comprehensive, customizable, and efficient modular RAG framework.}\quad 
\ours{} offers a highly modular setup at both the component and pipeline levels, featuring 5 core modules and 16 diverse RAG subcomponents that can be independently integrated or combined into pipelines. Additionally, we provide 9 standardized RAG processes and auxiliary scripts for tasks such as downloading and chunking Wikipedia for corpus construction, building retrieval indexes, and preparing retrieval results, resulting in an efficient and user-friendly end-to-end RAG framework.

\textbf{Pre-implemented advanced RAG algorithms.}\quad
To our knowledge, \ours{} has provided the most comprehensive implementation of existing work, including 16 advanced RAG algorithms, such as Self-RAG~\cite{asai2024selfrag} and FLARE~\cite{jiang2023active}, covering sequential, conditional, branching, and loop RAG categories. These methods are evaluated within a unified framework, and benchmark reports are available, supporting transparent evaluation and comparison.
More approaches will continue to be incorporated into our library.

\textbf{Support for multi-modal RAG scenarios.}
\ours{} supports comprehensive multi-modal RAG deployments by integrating mainstream multi-modal large language models (MLLMs) like Qwen~\cite{qwen2vl}, InternVL~\cite{internvl} and LLaVA~\cite{liu2023llava,liu2024llavanext}, along with various CLIP-based retrievers~\cite{CLIP}. The framework provides researchers with extensive technical support across both text-only and multi-modal scenarios, equipped with multiple widely-used MRAG benchmark datasets for systematic evaluation.

\begin{table}[!t]
\caption{The overall comparison between \ours{} and other toolkits.}
\small
\centering
\setlength{\tabcolsep}{11pt}{
\begin{tabular}{lccccc}
\toprule
\textbf{Toolkit} & \makecell[c]{\textbf{Automatic}\\ \textbf{Evaluation}} & \makecell[c]{\textbf{Multimodal}} & \makecell[c]{\textbf{Corpus}\\ \textbf{Helper}} & \makecell[c]{\textbf{\# Provided}\\ \textbf{Dataset}} & \makecell[c]{\textbf{\# Support}\\ \textbf{Methods}} \\ 
\midrule
Langchain~\cite{LangChain} & \xmark & \cmark & \cmark & - & 2 \\ 
LlamaIndex~\cite{LlamaIndex_2022} & \cmark & \cmark & \cmark & - & 2 \\ 
Haystack~\cite{Haystack_2019} & \cmark & \xmark & \xmark & - & - \\ 
FastRAG~\cite{fastRAG_2023} & \xmark & \xmark & \xmark & 2 & 1 \\ 
LocalRQA~\cite{localrqa} & \cmark & \xmark & \xmark & 3 & - \\ 
AutoRAG~\cite{autorag} & \cmark & \xmark & \xmark & 4 & 2 \\ 
RAGLab~\cite{raglab} & \cmark & \xmark & \cmark & 10 & 6 \\ 
\textbf{\ours{}} (ours) & \cmark & \cmark & \cmark & \textbf{38} & \textbf{16} \\ 
\bottomrule
\end{tabular}
}
\vspace{-2em}
\label{tab:main}
\end{table}

\textbf{Comprehensive benchmark datasets.}\quad
To improve the consistency and utility of datasets in RAG research, we have collected 38 commonly used datasets and standardized their formats. Some of these datasets, such as WikiAsp~\cite{wikiasp} and NarrativeQA~\cite{narrativeqa}, have undergone specific adjustments for RAG scenarios to ensure consistency. These datasets are readily available on the HuggingFace platform, facilitating easy access and application.


\textbf{Visual web interface for RAG experimentation.}
\ours{} features an intuitive web interface that visualizes the complete RAG pipeline. Users can inspect intermediate results at each step, from retrieval to answer generation, while performing one-click parameter tuning and automatic benchmark evaluation. The interface enables seamless corpus loading, real-time component visualization, and comprehensive pipeline assessment, making RAG experimentation more transparent and efficient.

\section{Background and Related work}

\subsection{Retrieval-Augmented Generation (RAG)}
Hallucinations and factual inaccuracies present significant challenges in existing LLMs~\cite{bang2023hallucination,search_o1, metarag,autoif}. To address these problems, RAG has been introduced. The typical RAG architecture includes a retriever and a generator (\textit{e.g.}, LLMs). The retriever retrieves relevant passages from an external knowledge base in response to a user’s query~\cite{unigen_xiaoxi, followrag_dgt, session_yiruo}, and then the generator uses these passages as part of the input for generation. By leveraging external knowledge, the quality of the generation can be significantly enhanced~\cite{trustworthy_survey, RetroLLM, htmlrag, conversationalsearch_survey}. As research in the RAG field progresses, various additional components have been introduced to further improve the effectiveness and robustness of RAG systems~\cite{rag_survey_gao,generative_survey_xiaoxi, slimplm_jiejun}. This motivates us to develop an accessible, user-friendly toolkit designed specifically to support and facilitate RAG research.


\subsection{RAG Toolkits}
RAG typically involves complex components and preliminary engineering tasks, such as corpus construction and generator setup. Due to the absence of a dedicated RAG library for research, existing open-source codes often use custom implementations with specific environment configurations, hindering code reusability and adaptation to new settings.

Recent developments have seen the introduction of several open-source RAG toolkits such as Langchain~\cite{LangChain}, LlamaIndex~\cite{LlamaIndex_2022}, and Haystack~\cite{Haystack_2019}. These libraries provide advanced APIs facilitate interactions with LLMs, such as vector databases and embedding models, simplifying the execution of RAG processes.  However, these toolkits generally do not cater to the needs of the research community. They often lack comprehensive implementations of existing RAG methods, do not provide access to commonly used retrieval corpora, and are typically heavy and overly encapsulated, which obscures details and complicates customization.

To address these challenges, lighter and more adaptable RAG toolkits have emerged. For example, FastRAG~\cite{fastRAG_2023} builds upon Haystack’s API to offer optimized support with a select range of methods and datasets. LocalRQA~\cite{localrqa} focuses on the training stage of the RAG process, providing training scripts for various components (such as retrievers and generators). AutoRAG~\cite{autorag} adopts a modular approach by treating each RAG component as a node that connects to form the overall process, but it lacks the implementation of existing RAG methods and their evaluations. In contrast, our \ours{} toolkit provides a wide array of RAG components and pre-implemented methods, allowing for straightforward replication of existing RAG studies in various settings with a few lines of code. Furthermore, we provide a rich resource pool, including a large number of processed datasets and scripts for obtaining and pre-processing widely used corpora, significantly reducing the preparatory time required for researchers.

\section{The Toolkit: \ours{}}
\begin{figure*}[t]
    \centering
    \includegraphics[width=0.95\textwidth]{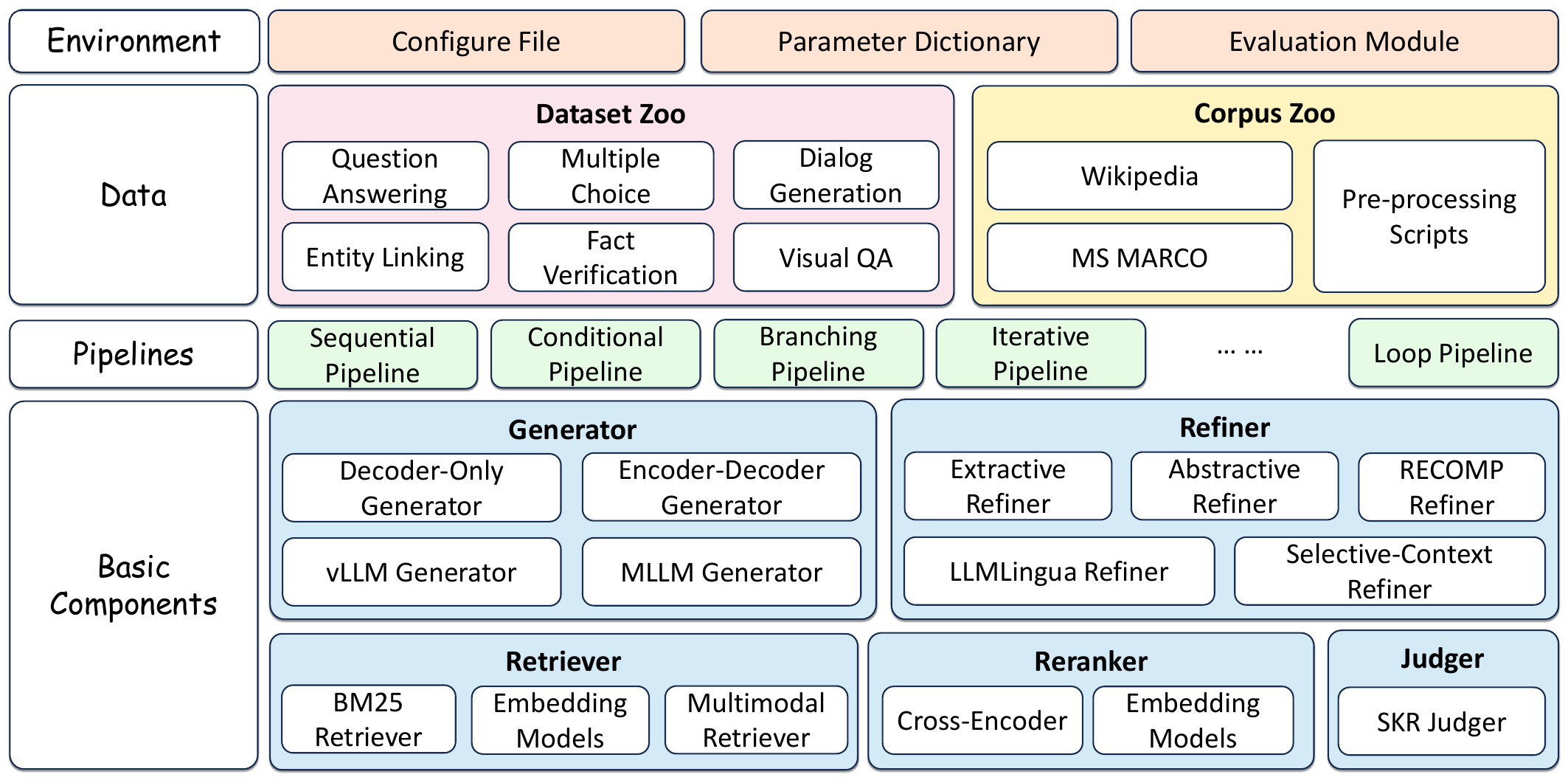}
    \caption{An overview of \ours{}.}
    \label{fig:main}
\end{figure*}

\ours{} is designed primarily to support research in RAG, although its capabilities are not confined to this area alone. As illustrated in Figure~\ref{fig:main}, \ours{} comprises three hierarchical modules: the environment module, the component module, and the pipeline module. The environment module is fundamental to the toolkit, providing essential resources such as datasets, hyperparameters, and evaluation metrics, for experiments. Building upon the environment module, the component module consists of various RAG components, each tailored for specific functions (\textit{e.g.}, retrieval and generation). The pipeline module integrates these components into a complete RAG process. In this paper, we will introduce the component and pipeline modules. Additional details are available in the documentation of our library.

\subsection{Component Module}
\label{subsection:component}
\ours{} is organized into five main components, each designed to function autonomously or within a combined application, enhancing both flexibility and specificity in the RAG process.

\textbf{Judger} functions as a preliminary component that determines whether a query needs retrieval~\cite{skr_2023,adaptiverag, dparag_dgt}. Given the limited studies in this domain, we implement a judger based on SKR~\cite{skr_2023}, which utilizes LLM self-knowledge data to determine the necessity of retrieval.

\textbf{Retriever} implementations are extensively covered by our toolkit. For sparse retrieval, we integrate the Pyserini library~\cite{Lin_etal_SIGIR2021_Pyserini} to support the use of BM25~\cite{bm25}. For dense retrieval, we consider various BERT-based embedding models such as DPR~\cite{dpr_2020}, E5~\cite{wang2022e5}, and BGE~\cite{bge_embedding}, along with T5-based models such as ANCE~\cite{ance_2021}. We employ FAISS~\cite{douze2024faiss,faiss_gpu} for efficient vector database operations and utilize the HuggingFace datasets library to enhance corpus loading speed. Additionally, \ours{} includes a ``retrieval cache'' feature, allowing for the reuse of retrieval results and supporting custom formats from non-open source retrievers.

\textbf{Reranker} aims at refining the order of retrieved results to improve retrieval accuracy. \ours{} supports many cross-encoder models, such as bge-reranker~\cite{bge_embedding} and jina-reranker, and facilitates the use of bi-encoder models such as E5 for scenarios using embedding models for reranking. Rerankers can be seamlessly integrated with any retriever through a decorator, enabling simple and flexible combinations.

\textbf{Refiner} processes input text to optimize it for generation by reducing token usage and noise.
We have implemented three types of refiners: extractive, abstractive, perplexity-based~\cite{jiang-etal-2023-llmlingua,jiang-etal-2023-longllmlingua,li2023selectivecontext}. Each type employs different methods to handle retrieved passages, such as semantic extraction or summarization~\cite{bider_jiajie}, utilizing both dedicated models like RECOMP~\cite{xu2023recomp} and general models available on HuggingFace.

\textbf{Generator} is the final component in the RAG process. We integrate two advanced LLM acceleration libraries, vLLM~\cite{vllm_2023} and FastChat~\cite{fastchat}, supporting a wide range of LLMs. Furthermore, we provide the native interface of the Transformers library~\cite{wolf-etal-2020-transformers} to enhance robustness. The module also includes encoder-decoder models like Flan-T5~\cite{flan} and utilizes fusion-in-decoder techniques~\cite{fid_2021} to enhance processing efficiency with retrieved content.

\begin{table*}
\caption{Summary of datasets. The sample size of each dataset and the knowledge source of the answer are listed as references. ``-'' indicates that the knowledge source is commonsense. The $\ast$ symbol represents that the task of this dataset has been modified to fit RAG settings.}
\small
\centering
\setlength{\tabcolsep}{4pt}{
\begin{tabular}{cllrrr}
\toprule
    \textbf{Task} & \textbf{Dataset Name} & \textbf{Knowledge Source} & \textbf{\# Train} & \textbf{\# Dev} & \textbf{\# Test} \\ 
\midrule
    \multirow{14}{*}{{QA}} & NQ~\cite{naturalquestion} & Wiki & 79,168 & 8,757 & 3,610 \\ 
    & TriviaQA~\cite{triviaqa} & Wiki \& Web & 78,785 & 8,837 & 11,313 \\ 
    & PopQA~\cite{popqa} & Wiki & - & - & 14,267 \\ 
    & SQuAD~\cite{squad} & Wiki & 87,599 & 10,570 & - \\ 
    & MSMARCO-QA~\cite{msmarco} & Web & 808,731 & 101,093 & - \\ 
    & NarrativeQA~\cite{narrativeqa} & Books, movie scripts & 32,747 & 3,461 & 10,557 \\ 
    & WikiQA~\cite{wikiqa} & Wiki & 20,360 & 2,733 & 6,165 \\ 
    & WebQuestions~\cite{webquestions} & Google Freebase & 3,778 & - & 2,032 \\ 
    & AmbigQA~\cite{ambigqa,naturalquestion} & Wiki & 10,036 & 2,002 & - \\ 
    & SIQA~\cite{siqa} & - & 33,410 & 1,954 & - \\ 
    & CommonsenseQA~\cite{commonsenseqa} & - & 9,741 & 1,221 & - \\ 
    & BoolQ~\cite{boolq} & Wiki & 9,427 & 3,270 & - \\ 
    & PIQA~\cite{piqa} & - & 16,113 & 1,838 & - \\ 
    & Fermi~\cite{fermi} & Wiki & 8,000 & 1,000 & 1,000 \\ 
\midrule
    \multirow{5}{*}{Multi-Hop QA} &HotpotQA~\cite{hotpotqa} & Wiki & 90,447 & 7,405 & - \\ 
    &2WikiMultiHopQA~\cite{2wikimultihop} & Wiki & 15,000 & 12,576 & - \\ 
    &Musique~\cite{musique} & Wiki & 19,938 & 2,417 & - \\ 
    &Bamboogle~\cite{selfask_2023} & Wiki & - & - & 125 \\ 
    &StrategyQA~\cite{strategyqa} & Wiki & 2,290 & - & - \\ \midrule
    \multirow{3}{*}{Long-Form QA}&ASQA~\cite{asqa} & Wiki & 4,353 & 948 & - \\ 
    &ELI5~\cite{eli5} & Reddit & 272,634 & 1,507 & - \\ 
    &WikiPassageQA~\cite{wikipassageqa} & Wiki & 3,332 & 417 & 416 \\ \midrule 
    \multirow{6}{*}{Multiple-Choice} &MMLU~\cite{mmlu,mmlu_ethics} & - & 99,842 & 1,531 & 14,042 \\ 
    &TruthfulQA~\cite{truthfulqa} & Wiki & - & 817 & - \\ 
    &HellaSwag~\cite{hellaswag} & ActivityNet & 39,905 & 10,042 & - \\ 
    &ARC~\cite{arc_challenge} & - & 3,370 & 869 & 3,548 \\ 
    &OpenBookQA~\cite{OpenBookQA2018} & - & 4,957 & 500 & 500 \\ 
    &QuaRTz~\cite{quartz} & - & 2,696 & 384 & 784 \\ \midrule
    \multirow{2}{*}{Entity linking} &AIDA CoNLL-YAGO~\cite{AIDA_CONLL,kilt_2021} &  Wiki \& Freebase & 18,395 & 4,784 & - \\ 
    &WNED~\cite{wned,kilt_2021} & Wiki & - & 8,995 & - \\ \midrule
     \multirow{2}{*}{Slot Filling} &T-REx~\cite{trex,kilt_2021} & DBPedia & 2,284,168 & 5,000 & - \\ 
    &Zero-shot RE~\cite{zeroshotre,kilt_2021} & Wiki & 147,909 & 3,724 & - \\ \midrule
    {Fact Verification} &FEVER~\cite{fever,kilt_2021} & Wiki & 104,966 & 10,444 & - \\ \midrule
    {Dialog Generation}&WOW~\cite{dinan2018wizard,kilt_2021} & Wiki & 63,734 & 3,054 & - \\ \midrule
    \makecell[c]{Open-domain \\ Summarization$^\ast$} &WikiAsp~\cite{wikiasp} & Wiki & 300,636 & 37,046 & 37,368 \\ \midrule
    {In-domain QA}& DomainRAG~\cite{domainrag} & Web pages of RUC & - & - & 485 \\ \midrule
    \multirow{3}{*}{Vision QA}&Gaokao-MM~\cite{gaokaomm} & - & 519 & - & 127 \\ 
    &MultimodalQA~\cite{multimodalqa} & Wiki & 2,099 & 230 & - \\ 
    &MathVista~\cite{mathvista} & - & - & - & - \\ 
    \bottomrule
\end{tabular}
}
\label{tab:dataset}
\end{table*}

\subsection{Pipeline Module}
Building on the diverse components, \ours{} allows the decoupling of the algorithmic flow of the RAG process from the specific implementations of each component. 
In constructing the pipeline, users only need to determine the required components for the RAG process and the logic of data flow between these components. 
The pipeline will automatically executes the corresponding RAG process and provides both the intermediate results and evaluation results.

To systematically execute the operational logic of various RAG tasks, we conduct an in-depth analysis of RAG-related literature. Following a recent survey on RAG~\cite{rag_survey_gao}, we identify four primary types of RAG process flows: Sequential, Branching, Conditional, and Loop. 

\textbf{Sequential Pipeline} implements a linear execution path, typically represented as ``query $\rightarrow$ retriever $\rightarrow$ post-retrieval (reranker, refiner) $\rightarrow$ generator''. Once the user has configured their settings, the library automatically loads the necessary components along with their corresponding process logic.

\textbf{Branching Pipeline} executes multiple paths in parallel for a single query (often one path per retrieved passage) and merges the results from all paths to form the final output. Currently, \ours{} supports two advanced branching methods: REPLUG pipeline~\cite{replug} and SuRe pipeline~\cite{sure_2024}. The REPLUG pipeline processes each retrieved passage in parallel and combines the generation probabilities from all passages to produce the final answer. The SuRe pipeline generates a candidate answer from each retrieved passage and then ranks all candidate answers. 

\textbf{Conditional Pipeline} utilizes a judger to direct the query into different execution paths based on predefined criteria. Queries requiring retrieval follow the standard sequential process, while others bypass retrieval and proceed directly to generation. \ours{} provides utility functions to split and merge the input dataset based on the judger's decisions, ensuring batch processing and enhanced pipeline efficiency. Additionally, the conditional pipeline supports integration with various types of pipelines, enabling dynamic execution based on judger's results.

\textbf{Loop Pipeline} involves complex interactions between retrieval and generation processes, often containing multiple cycles of retrieval and generation. Compared to the previous ones, this pipeline is more flexible, and thus can often yield better performance. \ours{} currently supports four typical methods, including Iterative~\cite{iterretgen_2023,itrg2023}, Self-Ask~\cite{selfask_2023}, Self-RAG~\cite{asai2024selfrag}, and FLARE~\cite{jiang2023active}.

\begin{table*}
\caption{The benchmarking results. \textit{Optimize component} represents the primary component optimized by the method, while \textit{flow} indicates optimization of the entire RAG process. Methods marked with $\ast$ denote the use of a trained generator.}
\small
\centering
\setlength{\tabcolsep}{2.3pt}{
\begin{tabular}{lcccccccc}
\toprule
& Optimize & Pipeline& NQ & TriviaQA & HotpotQA & 2Wiki & PopQA & WebQA \\
\multirow{-2}{*}{Method} &  component & type & (EM) & (EM) & (F1) & (F1) & (F1) & (EM)\\
\midrule
Naive Generation & - &Sequential & 22.6 & 55.7 & 28.4 & 33.9 & 21.7 & 18.8 \\
Standard RAG & - & Sequential & 35.1 & 58.8 & 35.3 & 21.0 & 36.7 & 15.7 \\
AAR~\cite{aar_retriever_2023} & Retriever &Sequential & 30.1 & 56.8 & 33.4 & 19.8 & 36.1 & 16.1 \\
LongLLMLingua~\cite{jiang-etal-2023-longllmlingua} & Refiner & Sequential & 32.2 & 59.2 & 37.5 & 25.0 & 38.7 & 17.5 \\
RECOMP-abstractive~\cite{xu2023recomp} & Refiner & Sequential & 33.1 & 56.4 & 37.5 & 32.4 & 39.9 & 20.2 \\
Selective-Context~\cite{li2023selectivecontext} & Refiner & Sequential & 30.5 & 55.6 & 34.4 & 18.5 & 33.5 & 17.3 \\
Trace ~\cite{trace} & Refiner & Sequential & 30.7 & 50.2 & 34.0 & 15.5 & 37.4 & 19.9 \\
Spring ~\cite{spring} & Generator & Sequential & 37.9 & 64.6 & 42.6 & 37.3 & 54.8 & 27.7 \\
$\text{Ret-Robust}^\ast$~\cite{retrobust_2023} & Generator &Sequential & 42.9 & 68.2 & 35.8 & 43.4 & 57.2 & 33.7  \\
SuRe~\cite{sure_2024} & Flow & Branching & 37.1 & 53.2 & 33.4 & 20.6 & 48.1 & 24.2  \\
REPLUG~\cite{replug} &  Generator & Branching & 28.9 & 57.7 & 31.2 & 21.1 & 27.8 & 20.2 \\
SKR~\cite{skr_2023} & Judger & Conditional & 33.2 & 56.0 & 32.4 & 23.4 & 31.7 & 17.0 \\
Adaptive-RAG ~\cite{adaptiverag} & Judger & Conditional & 35.1 & 56.6 & 39.1 & 28.4 & 40.4 & 16.0 \\
$\text{Self-RAG}^\ast$~\cite{asai2024selfrag} & Flow & Loop & 36.4 & 38.2 & 29.6 & 25.1 & 32.7 & 21.9 \\
FLARE~\cite{jiang2023active} & Flow & Loop & 22.5 & 55.8 & 28.0 & 33.9 & 20.7 & 20.2\\
Iter-RetGen~\cite{iterretgen_2023}, ITRG~\cite{itrg2023} & Flow & Loop & 36.8 & 60.1 & 38.3 & 21.6 & 37.9 & 18.2 \\
IRCoT~\cite{ircot} & Flow & Loop & 33.3 & 56.9 & 41.5 & 32.4 & 45.6 & 20.7 \\
RQRAG~\cite{rqrag} & Flow & Loop & 32.6 & 52.5 & 33.5 & 35.8 & 46.4 & 26.2 \\
\bottomrule
\end{tabular}
}
\label{tab:main_result}
\end{table*}

\subsection{Datasets}
\subsubsection{Datasets for RAG}
We collect and pre-process 32 benchmark datasets, covering a broad spectrum utilized in existing studies on RAG. The statistics of these datasets are shown in Table~\ref{tab:dataset}. Each dataset has been standardized into a unified JSONL format, comprising four fields per item: ID, question, golden answer, and metadata. For multiple-choice datasets such as MMLU~\cite{mmlu,mmlu_ethics}, an additional ``choices'' field is provided as options. These processed datasets are readily accessible on HuggingFace. Further details on dataset processing are available in Appendix~\ref{sec:datasets}.

In addition to the datasets, we provide a variety of dataset filtering tools that enable users to tailor the dataset according to their needs. For instance, user can select a certain number of samples, either randomly or sequentially, for evaluation, or filter subsets based on the dataset's metadata. These methods are unified within a dataset loading function, providing a consistent user interface. Customized filtering functions can also be implemented by users.

\subsubsection{Retrieval Corpus}
The corpus used for retrieval, also known as the knowledge base, is also crucial for preparing experiments. Typically used are the Wikipedia passages and MS MARCO passages. 

\textbf{Wikipedia passages}: The Wikipedia passages comprise a collection of passages from English Wikipedia pages, widely used as knowledge sources in datasets like KILT~\cite{kilt_2021}. The process of acquiring the Wikipedia corpus includes downloading snapshots in XML format, cleaning the text of HTML tags, extracting textual content, and segmenting it into passages suitable for retrieval.

In research, various versions of Wikipedia are often used, increasing the difficulty of reproduction.
We provide easy-to-use scripts for automatically downloading and pre-processing any required version of Wikipedia. Additionally, we provide various chunking functions to support customized segmentation, enabling alignment with standard or previously used corpora. Our toolkit includes the widely used Wikipedia dump from DPR~\cite{dpr_2020} dated December 20, 2018, as a fundamental resource.

\textbf{MS MARCO passages}~\cite{msmarco}: This dataset consists of 8.8 million passages sourced from Bing search engine. Compared to the Wikipedia dump, it contains fewer passages and has undergone pre-processing. This dataset has already been released on HuggingFace, so we provide direct links in our library for easy access.

\subsection{Evaluation Metrics}
\ours{} supports several commonly used evaluation metrics to measure the quality of the RAG process. Depending on the subject of evaluation, these metrics can be categorized into retrieval-aspect metrics and generation-aspect metrics.

\textbf{Retrieval-aspect metrics}: \ours{} supports four metrics including recall@$k$, precision@$k$, F1@$k$, and mean average precision (MAP) to evaluate retrieval quality. Different from evaluation in standalone retrieval systems, the passages retrieved in the RAG process often lack golden labels (\textit{e.g.}, related or unrelated tags). Therefore, we consider the presence of the golden answer within retrieved passages as an indicator of relevance. Other metrics can be implemented by inheriting existing metrics and modifying the calculation methods.

\textbf{Generation-aspect metrics}: To evaluate the quality of generation, \ours{} supports five metrics: token-level F1 score, exact match, accuracy, BLEU~\cite{bleu}, and ROUGE-L~\cite{lin-2004-rouge}. Moreover, \ours{} also evaluates the number of tokens used in generation to analyze cost-effectiveness.

To accommodate custom evaluation metrics, \ours{} provides a metric template for users. As our library automatically saves intermediate results, users can conveniently evaluate results from intermediate components, such as analyzing token usage pre- and post-refinement or comparing precision across multiple retrieval rounds.

\section{Textual Experimental Result and Discussion}
To validate the effectiveness of \ours{}, we conduct a series of experiments for providing reproducible benchmarking results and facilitating further exploration. In our main experiment, by default, we employ the latest \texttt{LLaMA-3-8B-instruct}~\cite{llama3modelcard} model as the generator and \texttt{E5-base-v2} as the retriever. Some methods may fine-tune their own models in different RAG components, and we provide the details of them in Appendix~\ref{sec:methods}. 

\textbf{Methods.}\quad We evaluate the performance of all supported RAG methods. These methods are categorized based on the RAG component they primarily aim to optimize: AAR~\cite{aar_retriever_2023} aims at optimizing the retriever; LongLLMLingua~\cite{jiang-etal-2023-longllmlingua}, RECOMP~\cite{xu2023recomp}, Selective-Context~\cite{li2023selectivecontext} and Trace~\cite{trace} focus on refining and compressing the input; Ret-Robust~\cite{retrobust_2023}, Spring~\cite{spring} and REPLUG~\cite{replug} aim to enhance the generator and its decoding approaches; SKR~\cite{skr_2023}, Adaptive-RAG~\cite{adaptiverag} introduce the judger to decide the necessity of retrieval for a query; SuRe~\cite{sure_2024}, Self-RAG~\cite{asai2024selfrag}, FLARE~\cite{jiang2023active}, Iter-RetGen~\cite{iterretgen_2023}, RQRAG~\cite{rqrag} and ITRG~\cite{itrg2023} optimize the entire RAG flow, including using multi-round retrieval and generation processes.

\subsection{Benchmarking Results}
The experimental results are shown in Table~\ref{tab:main_result}. Overall, RAG methods significantly outperform the direct generation baseline, which clearly demonstrates the benefits of incorporating external knowledge into the generation process. We further have the following observations: 
(1) Standard RAG, with advanced retrievers and generators, is a strong baseline, showing robust performance across six datasets. 
(2) AAR improves retrievers by fine-tuning the \texttt{contriever} model, achieving results comparable to the \texttt{E5} baseline on various datasets.
(3) All three methods employing refiners exhibit significant improvements, particularly on multi-hop datasets such as HotpotQA and 2WikiMultihopQA. This is potentially because complex problems result in less accurate passage retrieval, introducing more noise and highlighting the necessity for refiner optimization. 
(4) As for generator optimization method, Ret-Robust fine-tunes the \texttt{LLaMA-2-13B} model via LoRA~\cite{lora}, significantly enhancing generator's capability of understanding retrieved passages and outperforming other training-free approaches.
(5) The effectiveness of optimizing the RAG process varies depending on the dataset complexity. On simpler datasets such as NQ and TriviaQA, FLARE and Iter-RetGen perform comparably to, or slightly below, Standard RAG. In contrast, for more complex datasets requiring multi-step reasoning, such as HotpotQA, these methods demonstrate substantial improvements over the baseline. This indicates that adaptive retrieval methods are particularly advantageous for tackling complex problems, but they may introduce higher operational costs with limited benefits for simpler tasks.
\begin{figure}[t]
    
    \begin{minipage}[t]{0.5\linewidth}       
        \centering
        \includegraphics[width=\textwidth]{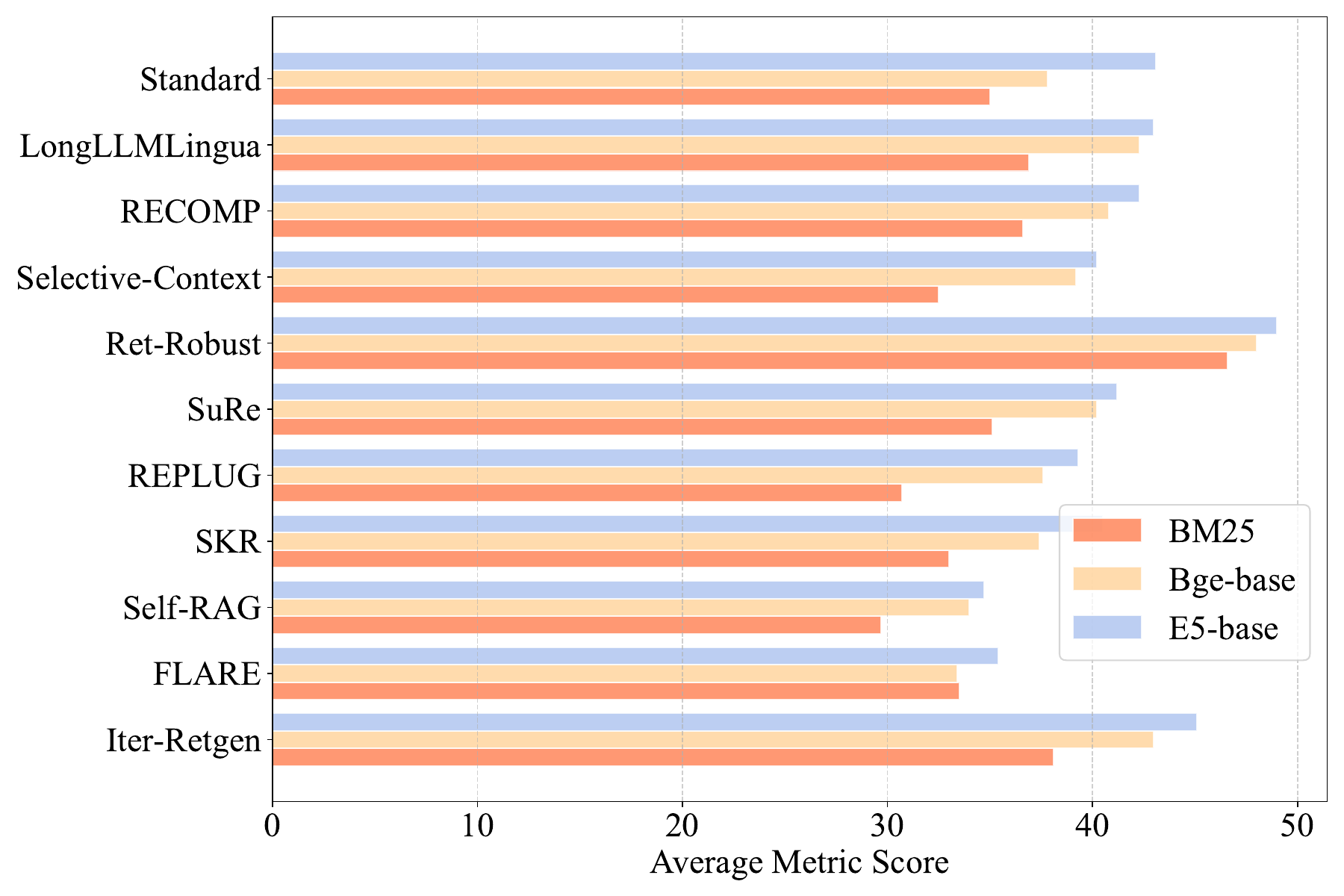}
    \end{minipage}%
    \begin{minipage}[t]{0.5\linewidth}
        \centering
        \includegraphics[width=\textwidth]{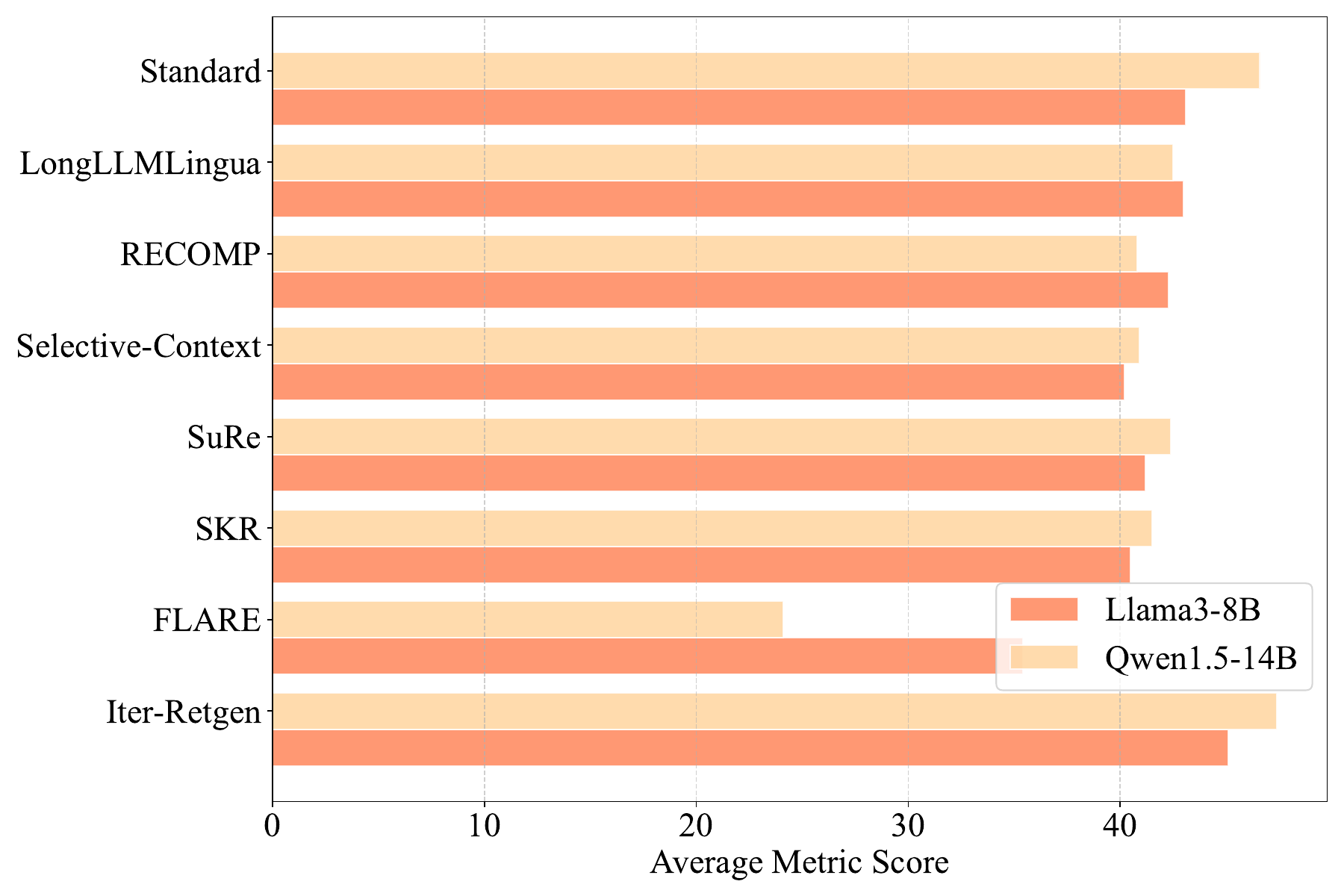}
    \end{minipage}
    \caption{The average results on three datasets(NQ, TriviaQA, HotpotQA) of baseline methods under different settings. \textbf{Left}: Results under three retrievers. \textbf{Right}: Results under two generator models with different parameter scale.}
    \label{fig:baseline}
\end{figure}

\subsection{Impact of Retrievers and Generators}
In RAG, the choice of retrievers and generators plays a crucial role in determining the final performance. Therefore, we conduct an experiment to explore their impact. Note that we do not include methods requiring specific retrievers or generators (\textit{e.g.}, Self-RAG requires trained models).

As shown in the left part of Figure~\ref{fig:baseline}, most methods are sensitive to retrieval quality. The performance gap between using the BM25 and E5 retriever can approach nearly 10\%. 
This gap is likely due to the presence of more noise in the retrieved passages of BM25, thereby disturbing the generation process with irrelevant information. In contrast, compression methods such as RECOMP show better robustness across various retrievers, suggesting that compression effectively mitigates the noise. Moreover, this robustness can be further enhanced by fine-tuning the generator. For example, Ret-Robust introduces a generator-specific training strategy that effectively minimizes the impact of irrelevant passages.

The influence of generators is also explored by using two models in different sizes (Qwen-1.5-14B and LLaMA-3-8B). Intriguingly, the larger model cannot consistently outperform the smaller one. For example, in methods such as FLARE and RECOMP, the smaller model yields better performance. Given that LLaMA-3-8B outperforms Qwen-1.5-14B in many public benchmarks, it suggests that the LLMs' RAG performance may be highly relevant to their general generation capabilities rather than their size. This observation highlights the complexity of LLMs' performance evaluation, suggesting that factors other than size, such as model architecture or training data quality, can also play significant roles.

\begin{figure*}[t]
    \centering
    \includegraphics[width=1.0\textwidth]{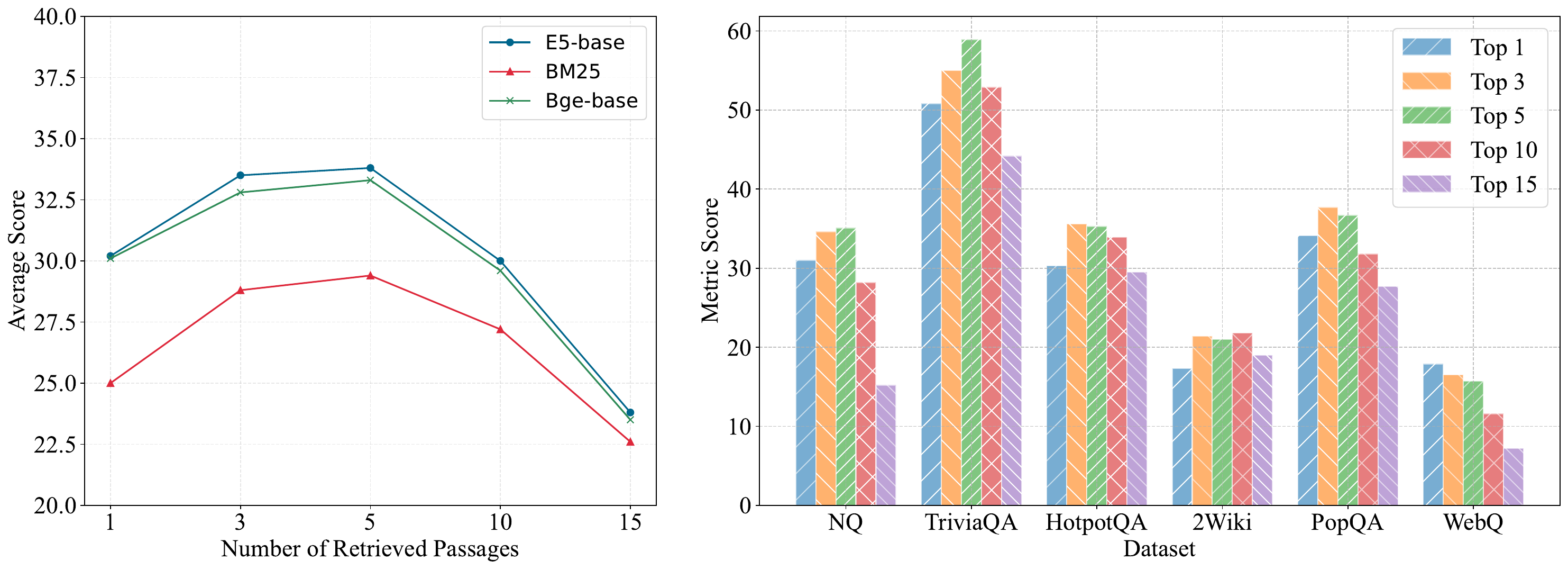}
    \caption{The results of standard RAG process under different number of retrieved passages and retrievers. \textbf{Left:} Average results on six datasets using three different retrievers with varying numbers of retrieved passages. \textbf{Right:} Individual results on six datasets using E5 as the retriever.}
    \label{fig:topk}
\end{figure*}

\subsection{Impact of Retrieval on RAG}
In RAG, both the quantity and quality of the retrieved passages significantly impact the final answer. However, existing studies often employ a predetermined retriever and a fixed number of retrieved passages, limiting further exploration. To comprehensively investigate the influence of the retrieval process on RAG results, we conduct a series of experiments. Figure~\ref{fig:topk} presents the results of using different numbers of retrieved passages. 

As shown in the left part, the best performance is achieved with approximately five retrieved passages. Both an excessive and insufficient number of passages lead to a significant drop in performance. This trend is consistent across various retrievers, including both dense and sparse methods. Moreover, the performance differences among various retrievers tend to diminish when a larger number of passages are used, suggesting an increase in noise within the retrieved content. In contrast, when only using the top-1 result, there is a substantial gap between dense methods (\textit{i.e.}, E5 and BGE) and BM25, indicating that with fewer passages, the quality of the retriever becomes a more critical factor in influencing the final performance.

In the right part, we analyze the impact of the number of retrieved passages on different datasets. For most datasets, the best performance is achieved with the top-5 retrieved results, suggesting that this may provide an ideal balance between the richness of information and potential noise. This finding highlights the importance of adjusting the retrieval count to optimize the effectiveness of the RAG process across various datasets.

\subsection{Impact of Corpus Chunking}
\begin{figure}[t]
    \centering
    \begin{minipage}[c]{0.48\linewidth}
        \centering
        \small
        \begin{tabular}{cccc}
        \toprule
        Chunk Size & Stride & \# Passages & Avg. Len \\
        \midrule
        6 sents  & 3 sents & 38.4 M & 127.7 \\
        6 sents & 6 sents & 23.4 M & 115.6 \\
        8 sents & 4 sents & 28.9 M & 163.5 \\
        10 sents & 5 sents & 23.3 M & 196.9 \\
        100 words & - & 30.1 M & 100.5 \\
        \bottomrule
        \end{tabular}
    \end{minipage}\hfill
    \begin{minipage}[c]{0.48\linewidth}
        \centering
        \includegraphics[width=\textwidth]{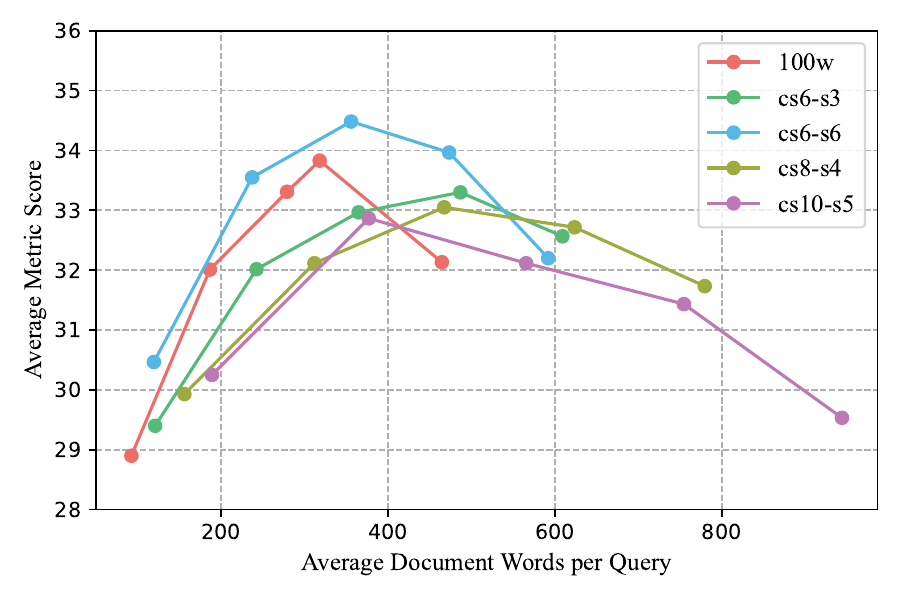}
        \label{fig:experimental_results}
    \end{minipage}
    \caption{{(Left)} Statistical overview of segmentation variants, where chunk size and stride are quantified in terms of \textit{sentences}, and average length denotes the average number of \textit{words} per passage. {(Right)} Experimental results of variants under different number of retrieval passages. }
\label{fig:compare_corpus}
\end{figure}

One key issue in RAG research is determining the optimal chunking method. Following previous studies in pre-processing Wikipedia~\cite{preprocessing,drqa}, we employ various chunking methods to segment the latest version of the Wikipedia dump into different variants and explore their corresponding performance.\footnote{Wikipedia Dump: \url{https://dumps.wikimedia.org/enwiki/}}
We experiment with four different chunk sizes: six sentences, eight sentences, ten sentences, and 100 words. Each configuration uses a stride set to half of the chunk size. Additionally, we explore a six-sentence chunk with a full chunk stride (non-overlapping) to evaluate the impact of stride variations on performance. To control the number of input words, we regulate the number of retrieved passages. The statistics are shown in the left part of Figure~\ref{fig:compare_corpus}.

The right part of Figure~\ref{fig:compare_corpus} illustrates the results. We can observe that optimal performance is typically achieved when each query corresponds to approximately 350-450 words of retrieved results, regardless of the chunk size. This suggests that with larger chunk sizes, a reduced number of passages may be beneficial. However, comparisons within the six-sentence chunks reveal that overlapping chunks tend to yield inferior performance. This highlights the nuanced effects that both chunk size and stride have on the efficacy of RAG.

\begin{table*}[ht]
\centering
\caption{The overall MRAG performance of different MLLM backbones.}
\label{tab:overall_mm_result}
\setlength\tabcolsep{4.6pt}
\fontsize{8.9pt}{11pt}\selectfont
\begin{tabular}{ccccccccc}
\toprule
\multirow{2}{*}{\textbf{MLLM}} & \multirow{2}{*}{\textbf{Retriever}} & \multicolumn{1}{c}{\textbf{Gaokao-MM}} & \multicolumn{3}{c}{\textbf{MultimodalQA}} & \multicolumn{3}{c}{\textbf{MathVista}} \\
\cmidrule(lr){3-3} \cmidrule(lr){4-6} \cmidrule(lr){7-9}
& & Acc & EM & F1 & Acc & EM & F1 & Acc \\
\midrule
Llava-7B & - & 0.142 & 0.317 & 0.401 & 0.739 & 0.262 & 0.162 & 0.389 \\
Llava-ov-7B & - & 0.374 & 0.743 & 0.752 & 0.761 & 0.581 & 0.457 & 0.649 \\
\midrule
\multirow{4}{*}{Qwen2-vl-7B} & - & 0.299 & 0.304 & 0.319 & 0.352 & 0.570 & 0.449 & 0.644 \\
& openai-clip & 0.339 & 0.348 & 0.351 & 0.352 & 0.444 & 0.344 & 0.535 \\
& jina-clip-v2 & 0.343 & 0.404 & 0.410 & 0.426 & 0.471 & 0.356 & 0.552 \\
& chinese-clip & 0.295 & 0.404 & 0.410 & 0.422 & 0.427 & 0.337 & 0.538 \\
\midrule
\multirow{4}{*}{Internvl2.5-8B} & - & 0.220 & 0.343 & 0.349 & 0.357 & 0.569 & 0.456 & 0.692 \\
& openai-clip & 0.287 & 0.439 & 0.447 & 0.452 & 0.450 & 0.347 & 0.580 \\
& jina-clip-v2 & 0.287 & 0.426 & 0.438 & 0.448 & 0.485 & 0.372 & 0.610 \\
& chinese-clip & 0.248 & 0.470 & 0.481 & 0.500 & 0.461 & 0.354 & 0.585 \\
\bottomrule
\end{tabular}
\end{table*}

\section{Multimodal Experimental Results and Analysis}
\subsection{Overall Results}
To further explore the performance of multimodal retrieval augmented  technologies in conjunction with different MLLM backbones, Table~\ref{tab:overall_mm_result} illustrartes the overall results of various MRAG configurations across three widely used VQA datasets. For each question, we retrieve the Top-1 relevant image-text pair from the corresponding retrieval corpus. In detail, we have identified the following key insights:

\textbf{MRAG delivers stable performance gains across MLLMs in knowledge-intensive tasks.} In knowledge-intensive multimodal QA datasets, multimodal retrieval knowledge significantly enhances the performance of MLLM bases, particularly the combination of Qwen2-VL and OpenAI-Clip, which provides an improvement of over 10 points in EM scores. Focusing on the Gaokao MM multidimensional reasoning assessment, multimodal retrieval knowledge also delivers stable enhancements, regardless of whether it involves Qwen2-VL-7B or InternVL-2.5-8B. This confirms that multimodal retrieval augmentation effectively supplements MLLMs in the domain of general knowledge reasoning.

\textbf{MRAG still struggles with multimodal mathematical reasoning.} However, when focusing on complex multimodal mathematical reasoning tasks, multimodal retrieval knowledge does not yield performance gains and instead results in noticeable negative effects. For instance, the combination of Clip and Qwen2 vL 7B even leads to a 10-point drop in accuracy. This observation aligns with findings from AR-MCTS~\citep{armcts}, indicating that in the specialized domain of mathematical reasoning, the in-domain multimodal retrieval augmentation method still exhibits significant limitations.

\begin{table*}[ht]
\centering

\caption{The overall result of different retrieved document quantity in MRAG under various retrieval types.}
\fontsize{8.9pt}{11pt}\selectfont
\renewcommand{\arraystretch}{0.9} 
\label{tab:mm_topk_result}
\begin{tabular}{cccccccc}
\toprule
\multirow{2}{*}{\textbf{MLLM}} & \multirow{2}{*}{\textbf{Retrieval Type}} & \multirow{2}{*}{\textbf{Retriever}} & \multirow{2}{*}{\textbf{Top-K}} & \textbf{GaoKao-MM} & \multicolumn{3}{c}{\textbf{Multimodal QA}} \\
\cmidrule(lr){5-5} \cmidrule(lr){6-8} 
& & & & Acc & EM & F1 \\ 
\midrule
\multirow{11}{*}{Qwen2-vl-7B} & \multirow{4}{*}{Multimodal} & \multirow{4}{*}{Clip} & 1 & 0.295  & 0.352  & 0.361  \\ 
 &  &  & 2 & 0.335  & 0.335  & 0.343  \\ 
 &  &  & 3 & 0.307  & 0.348  & 0.362  \\ 
 &  &  & 5 & 0.358  & 0.374  & 0.389  \\ 
\cmidrule(lr){2-8} 
 & \multirow{4}{*}{Text-only} & \multirow{4}{*}{BM25} & 1 & 0.35   & 0.335  & 0.344  \\ 
 &  &  & 2 & 0.307  & 0.343  & 0.36   \\ 
 &  &  & 3 & 0.307  & 0.357  & 0.376  \\ 
 &  &  & 5 & 0.358  & 0.365  & 0.391  \\ 
\cmidrule(lr){2-8} 
 & \multirow{3}{*}{Hybridmodal} & \multirow{3}{*}{BM25 + Clip} & 1 & 0.35  & 0.361  & 0.371  \\ 
 &  &  & 2 & 0.307  & 0.374  & 0.391  \\ 
 &  &  & 3 & 0.307  & 0.378  & 0.396  \\ 
\midrule
\multirow{11}{*}{Internvl2.5-8B} & \multirow{4}{*}{Multimodal} & \multirow{4}{*}{Clip} & 1 & 0.248  & 0.404  & 0.412  \\ 
 &  &  & 2 & 0.26   & 0.43   & 0.462  \\ 
 &  &  & 3 & 0.291  & 0.443  & 0.463  \\ 
 &  &  & 5 & 0.264  & 0.443  & 0.462  \\ 
\cmidrule(lr){2-8} 
 & \multirow{4}{*}{Text-only} & \multirow{4}{*}{BM25} & 1 & 0.287  & 0.4    & 0.408  \\ 
 &  &  & 2 & 0.272  & 0.37   & 0.381  \\ 
 &  &  & 3 & 0.26   & 0.383  & 0.396  \\ 
 &  &  & 5 & 0.248  & 0.4    & 0.417  \\ 
\cmidrule(lr){2-8} 
 & \multirow{3}{*}{Hybridmodal} & \multirow{3}{*}{BM25 + Clip} & 1 & 0.287  & 0.417  & 0.432  \\ 
 &  &  & 2 & 0.272  & 0.413  & 0.436  \\ 
 &  &  & 3 & 0.26   & 0.452  & 0.472  \\ 
\bottomrule
\end{tabular}
\end{table*}

\subsection{Impact of different Multimodal Retrievers}

To investigate the impact of various retrieval types, Table~\ref{tab:overall_mm_result} presents a comparison of three commonly used retrievers in MRAG: OpenAI-Clip\footnote{Model Card: \url{https://huggingface.co/openai/clip-vit-large-patch14}}, Jina-Clip-v2, and Chinese-Clip. 

\textbf{The language consistency of the retriever is not a decisive factor for MRAG.} In the Gaokao-MM evaluation, each retriever consistently offers improvements. Notably, Chinese-Clip does not achieve the highest gains, highlighting that the selection of multimodal retrievers should not solely depend on linguistic consistency. Similar findings emerge in the Multimodal QA, where all retrievers demonstrate stable enhancements, with the linguistically inconsistent Chinese-Clip showing the most substantial performance boost. 

However, in MathVista, the knowledge retrieved from different retrievers notably influences reasoning performance. We hypothesize that the retrieved knowledge may to some extent disrupt the coherence of problem-solving reasoning~\citep{searcho1,howabilities}.

\subsection{Influence of Retrieval Types \& Top-K Retrieved knowledge}

To comprehensively assess the impact of the number of multimodal retrieval documents on performance, we conducted evaluations using the top 1, 2, 3, and 5 documents in three different settings: text-only retrieval (BM25), multimodal retrieval (Clip), and hybridmodal retrieval (BM25+Clip). 

Our findings reveal that, in both the Gaokao-MM and MultimodalQA evaluation, utilizing Clip for multimodal retrieval generally results in more consistent improvements as the number of documents increases (from Top-1 to Top-3). Nonetheless, once the document count surpasses a certain threshold (3), MRAG's performance may become unstable. In the case of text-only and hybridmodal retrieval, MRAG continues to exhibit steady improvements with an increasing number of documents in MultimodalQA. Conversely, for Gaokao-MM, adding more documents does not consistently enhance performance, suggesting that complex reasoning datasets are particularly sensitive to the availability of knowledge.

\subsection{Effect of MLLM Parameter Size}

\begin{figure}[!t]
\centering
\includegraphics[width=0.9\linewidth]{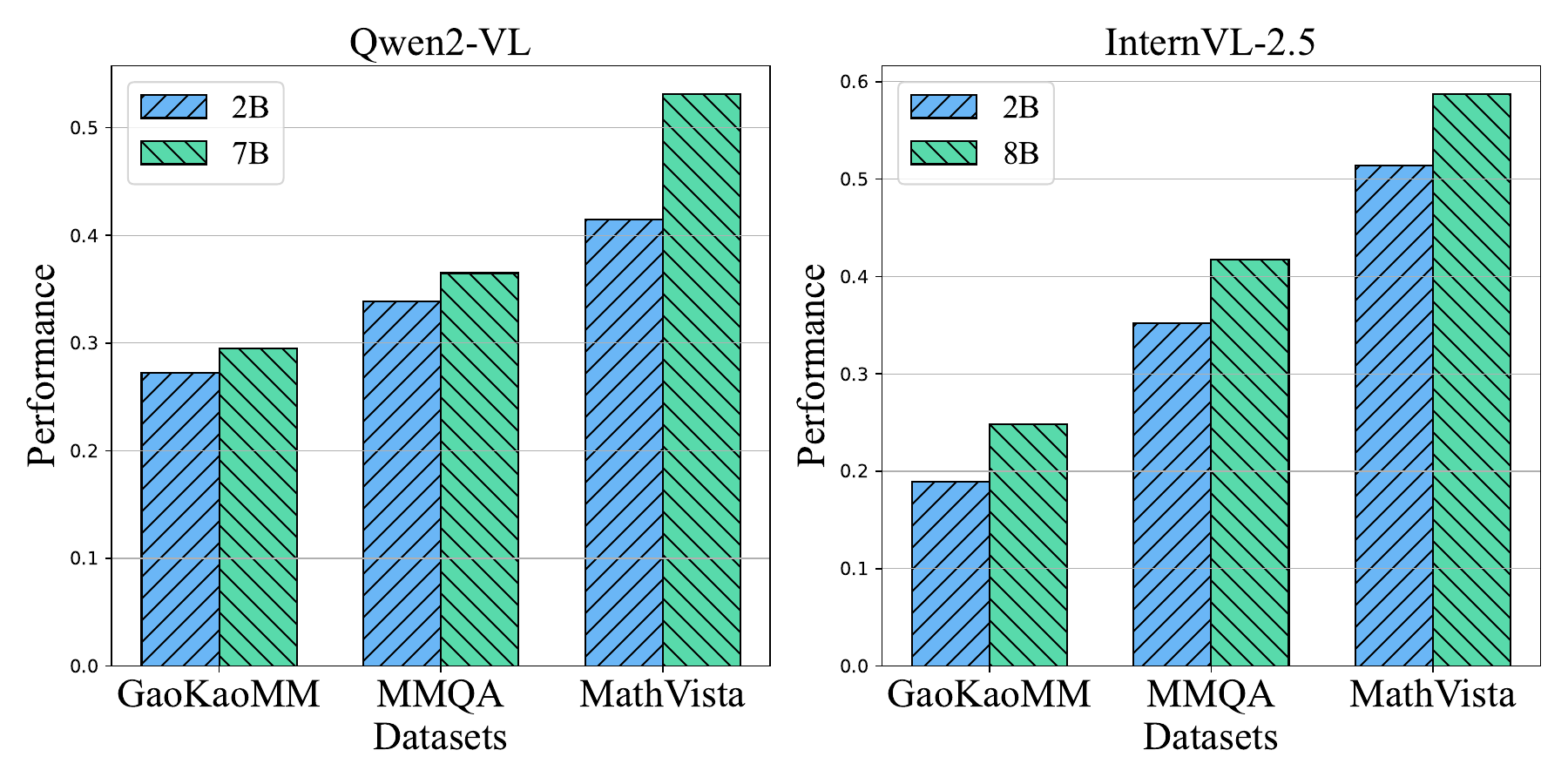}
\caption{Analysis of using MLLM backbone with difference param scales in MRAG.}
\label{fig:mm_params}
\end{figure}

In Figure~\ref{fig:mm_params}, we delve deeper into the influence of varying parameter sizes of MLLM backbones on MRAG's performance. Remarkably, both the Qwen2-VL and InternVL2.5 series exhibit consistent performance enhancements across all three datasets as the parameter size increases, especially in the math-specific reasoning benchmark MathVista. This reinforces the notion that MLLMs with larger parameters generally have superior abilities in harnessing retrieval knowledge.

\section{Conclusion}
In this work, we introduced FlashRAG, a new modular toolkit designed to address the challenges of replicability and the high development costs in RAG research. FlashRAG consists of a comprehensive collection of benchmark datasets, state-of-the-art RAG methods, utilities for corpus pre-processing, and a set of widely used evaluation metrics. This toolkit not only aids in the replication of existing RAG techniques but also supports the development of new approaches. Our experiments across multiple datasets have demonstrated the effectiveness of FlashRAG and have highlighted several critical factors for the successful development of RAG applications. By lowering the technical barriers and enhancing reproducibility, FlashRAG aims to accelerate the pace of innovation in the RAG domain, ultimately contributing to more robust and effective systems.


\bibliographystyle{unsrt}
\bibliography{Reference}
\clearpage

\appendix


\noindent\makebox[\linewidth]{\rule{\linewidth}{3.5pt}}
\begin{center}
	\bf{\Large Supplementary Material}
\end{center}
\noindent\makebox[\linewidth]{\rule{\linewidth}{1pt}}

This appendix is organized as follows.

\begin{itemize}
    \item In Section~\ref{sec:methods} (referred by Section 4), we introduce implementation details for our benchmarking experiments.
    \item In Section~\ref{sec:datasets} (referred by Section 3.3), we introduce the details in gathering and pre-processing datasets.
    \item In Section~\ref{sec:limitations}, we discuss the limitations of our toolkit.
\end{itemize}

\section{Implementation Details for Benchmarking Experiments}
\label{sec:methods}
In our work, we test all implemented methods under a unified setting. Users only need to download the corresponding model and fill in the configure to get the corresponding results using the script we provide.\footnote{Running Scripts: \url{https://github.com/RUC-NLPIR/FlashRAG/blob/main/examples/methods/run_exp.py}} This section details the implementation specifics of reproducing various algorithms using our toolkit, allowing users to effortlessly replicate our experimental results.

\subsection{Global Setting}

\textbf{Retriever Setting}: In our main experiments, we utilize \texttt{E5-base-v2} as the retriever, retrieving five passages per query.\footnote{Model Card: \url{https://huggingface.co/intfloat/e5-base-v2}} We use the DPR version of the Wikipedia December 2018 dataset as the retrieval corpus, which can be downloaded from our dataset page. 
In subsequent retrieval experiments, we employ both BM25 and \texttt{BGE-base-en-v1.5} as additional retrievers.\footnote{Model Card: \url{https://huggingface.co/BAAI/bge-base-en-v1.5}} The BM25 algorithms is implemented by Pyserini~\cite{pyserini}.
During index construction, the maximum padding length is set to 512. The maximum query padding length is set to 128 during retrieval. The batch size for retrieval is 1,024, with FP16 enabled. We employ the Faiss Flat index~\cite{faiss_gpu} for accuracy.

\textbf{Generator Setting}: We employ \texttt{LLaMA-3-8B-instruct} as the generator in our main experiment, with a maximum input length of 2048 and a maximum output length of 32.\footnote{\url{https://huggingface.co/meta-llama/Meta-Llama-3-8B-Instruct}} Inference is performed using the vLLM framework with greedy decoding during generation. In generator-related experiments, we employ Qwen-1.5-14B as additional generators. The experiments are conducted using four NVIDIA A100 80G GPUs.

\textbf{Prompt Setting}: We use a unified prompt to ensure fairness. Specifically, our system prompt is:

\begin{quote}
\texttt{Answer the question based on the given passage. Only give me the answer and do not output any other words. The following are given passages:\{retrieval passages\}}
\end{quote}
The retrieval passages are listed as:
\begin{quote}
    \texttt{Doc 1 (Title: \{title\}) \{content\}} \\
    \texttt{Doc 2 (Title: \{title\}) \{content\}}
\end{quote}

Our user prompt is:

\begin{quote}
\texttt{Question: \{question\}}
\end{quote}

These prompts are combined using the \texttt{tokenizer.apply\_chat\_template} function, serving as the final input to the generator model.

\subsection{Specific Settings for Different Methods}

In addition to the general settings mentioned above, each method often has its own configuration. We introduce it as follows:

\textbf{AAR}~\cite{aar_retriever_2023}: This work focuses on optimizing the retriever. In our experiments, we use the pre-trained retriever provided by the authors (\texttt{AAR-Contriever-KILT}).\footnote{\url{https://huggingface.co/OpenMatch/AAR-Contriever-KILT}} FlashRAG also supports using AAR-ANCE as the retriever.

\textbf{LLMLingua}~\cite{jiang-etal-2023-llmlingua, jiang-etal-2023-longllmlingua}: In this method, we use \texttt{LLaMA-2-7B} to compute perplexity and LongLLMLingua as the compressor, with a compression rate set to $0.55$. Other parameters are set to default values. Different from the original LongLLMLingua example, we only use the retrieved text as input to the refiner rather tan the entire prompt. This is because we find that the LLaMA-3's prompt requires special tokens, and using the original setting caused these tokens to be omitted, resulting in degraded performance.

\textbf{RECOMP}~\cite{xu2023recomp}: We use the abstractive model provided by RECOMP for our experiments.\footnote{\url{https://huggingface.co/fangyuan}} For the NQ, TQA, and HotpotQA datasets, we use the corresponding models. For the remaining datasets, there are no trained checkpoints available. Therefore, we use the HotpotQA checkpoint for 2WikiMultihopQA, and the NQ checkpoint for PopQA and WebQuestions. The maximum input length for the refiner is set to 1024, and the maximum output length is set to 512.

\textbf{Selective-Context}~\cite{li2023selectivecontext}: We use GPT-2 to compute perplexity and set the compression rate to 0.5. Similar to LongLLMLingua, we use the retrieved passages as input to the refiner.

\textbf{Ret-Robust}~\cite{retrobust_2023}: This method focuses on optimizing the generative model. It is trained with the Self-Ask prompt method. The authors provided the LoRA models trained on NQ and 2WikiMultihopQA.\footnote{\url{https://huggingface.co/Ori/llama-2-13b-peft-nq-retrobust}} Consequently, we test using the \texttt{LLaMA-2-13B} model loaded with the corresponding LoRA parameters. As there is no trained model for HotpotQA, we use the LoRA parameters trained on 2WikiMultihopQA. For the remaining datasets, we use the LoRA parameters trained on NQ. We set the maximum interaction rounds to five and the maximum output tokens to 100. For HotpotQA and 2WikiMultihopQA, we disable the ``single\_hop'' setting to allow the process to automatically decompose complex queries into multiple iterations.

\textbf{SuRe}~\cite{sure_2024}: This method prompts the model to generate candidate answers and scores, and then ranks them to select the best one. To ensure consistency, we use the prompts provided in the original paper, which can be referenced alongside our code implementation.

\textbf{SKR}~\cite{skr_2023}: We implement the SKR-KNN method, which requires an encoder model and inference-time training data. Specifically, it identifies the most similar queries from the training data based on the input query, determining whether the input query needs retrieval. Our library includes the training data provided by the authors. The corresponding encoder model is released by the authors.\footnote{\url{https://huggingface.co/princeton-nlp/sup-simcse-bert-base-uncased}}

\textbf{Self-RAG}~\cite{asai2024selfrag}: We use the \texttt{LLaMA-2-7B} checkpoint provided by Self-RAG and set the maximum number of output tokens to 100 to ensure proper operation.\footnote{\url{https://huggingface.co/selfrag/selfrag_llama2_7b}} The temperature is set to 0, and top\_p is set to 1.

\section{Collecting Details of Various Datasets}
\label{sec:datasets}
To facilitate users in understanding the sources and composition of the datasets, we have outlined the collection sources and pre-processing methods for each dataset provided by our library. For datasets with multiple versions, we refer to existing studies and select the version most widely used by researchers.

For datasets containing multiple subsets, we merge all subsets and indicate the subset each data point belongs to in the corresponding metadata attribute. This allows users to conveniently load specific subsets using the loading functions provided in our toolkit. Each dataset is stored in a separate folder, with all splits named ``train'', ``dev'', or ``test'', and each file is in JSONL format. Splits without golden answers are excluded from our consideration. For special cases involving multiple files, such as MMLU, we provide detailed explanations later. All datasets have been processed into a unified format, with each data point containing the following fields:

\begin{itemize}
\item \texttt{id}: A unique identifier for the data point, composed of the split it belongs to and its corresponding position.
\item \texttt{question}: This field generally represents the input portion of each data point. For example, for QA tasks, this is the question; for fact verification tasks, this is the claim to be judged.
\item \texttt{golden answers}: A list containing the correct answers. Even if there is only one correct answer, it is still stored in a list to maintain a unified format.
\item \texttt{choices}: A list of options, which is only available in datasets with multiple-choice tasks. For these datasets, their golden answers are the index corresponding to the correct options wrapped in a list.
\item \texttt{metadata}: A dictionary containing additional information about the dataset (\textit{e.g.}, subset, annotation data, etc.). During the collecting process, useful information from the original datasets is stored in the metadata for convenient data filtering by users in subsequent tasks.
\end{itemize}

The collecting details for each dataset are listed below.

\subsection{QA Datasets}
\textbf{Natural Questions (NQ)}~\cite{naturalquestion}, \textbf{TriviaQA (TQA)}~\cite{triviaqa}, \textbf{WebQuestions (WebQ)}~\cite{webquestions}: These are the most commonly used QA datasets, each available in multiple versions. We utilize the DPR~\cite{dpr_2020} versions of these datasets. For NQ, the training set is partially split into a validation set based on NQ-Open~\cite{nqopen} (which includes only training and test sets). Compared to the original NQ dataset, this version has fewer examples as it excludes instances without golden annotations. For TQA, this version employs a re-split of the original training and test sets, similar to the approach used in Self-RAG~\cite{asai2024selfrag}.

\textbf{PopQA}~\cite{popqa}: This dataset is collected from the HuggingFace repository \url{https://huggingface.co/datasets/akariasai/PopQA}. We reformat the dataset, including re-encoding IDs, modifying golden answer fields, and setting metadata.

\textbf{Squad}~\cite{squad}: The dataset is sourced from the HuggingFace repository \url{https://huggingface.co/datasets/lhoestq/squad}, based on Squad v1.1. We merge the annotated context and title fields and remove redundant answer start positions.

\textbf{Fermi}~\cite{fermi}: This dataset is collected from the HuggingFace repository \url{https://huggingface.co/datasets/jeggers/fermi}. We merge the Fermi-real and Fermi-synth versions and label the type field in the metadata to indicate the source dataset.

\textbf{MS MARCO QA}~\cite{msmarco}: This dataset is released by Microsoft, which can be directly downloaded from the URL: \url{https://microsoft.github.io/MSMARCO-Question-Answering}. 

\textbf{NarrativeQA}~\cite{narrativeqa}: This dataset is released by DeepMind, which can be downloaded from the URL: \url{https://github.com/deepmind/narrativeqa}. The text field from the original data's \texttt{question} is used as the question, while the text field from the original data's \texttt{answers} is designated as the \texttt{golden\_answers}. Other field information is preserved in the metadata.

\textbf{SIQA}~\cite{siqa}: This dataset can be downloaded from the URL: \url{https://leaderboard.allenai.org/socialiqa/submissions/get-started}. The \texttt{answerA}, \texttt{answerB}, and \texttt{answerC} fields from the original data are combined into a list and used as the \texttt{golden\_answers}. Other field information is preserved in the metadata. 

\textbf{PIQA}~\cite{piqa}: This dataset is crafted for probing the physical commonsense capabilities of NLP models. We obtain it from \url{https://github.com/ybisk/ybisk.github.io/tree/master/piqa}. The original data comprises four fileds: \texttt{goal}, \texttt{sol1}, \texttt{sol2}, and \texttt{label}. The \texttt{label} indicates the correct solution, with ``0'' corresponding to \texttt{sol1} and ``1'' to \texttt{sol2}. We only retain the correct solution in our dataset, omitting the incorrect option and the \texttt{label} field.

\textbf{BoolQ}~\cite{boolq}: BoolQ is a dataset specifically designed for ``yes/no'' question answering, encompasses 15,942 examples. We obtain it from its official repository: \url{https://github.com/google-research-datasets/boolean-questions}. 

\textbf{AmbigQA}~\cite{ambigqa}: We obtain this dataset from its official repository: \url{https://nlp.cs.washington.edu/ambigqa/}. The \texttt{annotations} are relabeled as \texttt{golden\_answers}, and the \texttt{viewed\_doc\_titles}, \texttt{used\_queries}, \texttt{nq\_answer}, and \texttt{nq\_doc\_title} are integrated into the metadata. 

\textbf{CommonsenseQA}~\cite{commonsenseqa}: This dataset is a multiple-choice question answering dataset, necessitates a diverse array of commonsense knowledge to infer the correct answers. We obtain it from the official URL: \url{https://www.tau-nlp.sites.tau.ac.il/commonsenseqa}. The correct response from choices is preserved according to the \texttt{answerKey}, with the original choices, \texttt{answerKey}, and \texttt{question\_concept} are set in the metadata.

\textbf{WikiQA}~\cite{wikiqa}: WikiQA is an annotated resource of question-sentence pairs developed by Microsoft to enhance open-domain question answering research, originally includes \texttt{question}, \texttt{answer}, \texttt{document\_title}, and \texttt{label}. We obtain it from official URL: \url{https://aka.ms/WikiQA}. In our adaptation, \texttt{document\_title} and \texttt{label} are retained within the metadata. 

\subsection{MultihopQA Datasets}
\textbf{HotpotQA}~\cite{hotpotqa}: We collect the distractor version of the HotpotQA dataset from the official HuggingFace repository \url{https://huggingface.co/datasets/hotpot_qa}. We retain the question type, level, and all annotated supporting facts. The original answer strings are converted into single-element lists, with additional information integrated into the metadata.

\textbf{MuSiQue}~\cite{musique}: We download the MuSiQue-Ans version of the dataset from its official repository \url{https://github.com/StonyBrookNLP/musique}. Since the test set lacks golden answers, we excluded it from our collections.

\textbf{2WikiMultihopQA}~\cite{2wikimultihop}: We obtain the latest version of this dataset from its official repository. The original dataset includes a file recording aliases for some answer entities; we match these aliases for each entry and add them to the golden answers to ensure evaluation accuracy. Similar to HotpotQA, we place supporting facts and other additional information into the metadata.

\textbf{Bamboogle}~\cite{selfask_2023}: We collect and process this dataset from the official repository link \url{https://docs.google.com/spreadsheets/d/1jwcsA5kE4TObr9YHn9Gc-wQHYjTbLhDGx6tmIzMhl_U/edit#gid=0}.

\subsection{Long-Form QA Datasets}
\textbf{ASQA}~\cite{asqa}: The dataset is collected from the HuggingFace repository \url{https://huggingface.co/datasets/din0s/asqa}. We reprocess unnecessary structures and selected the long answer part from the annotations as the final golden answer, with other annotated content and QA pairs placed into the metadata.

\textbf{ELI5}~\cite{eli5}: The dataset is collected from the KILT benchmark~\cite{kilt_2021}. Due to the lack of correct answers in the test set, we only retain the training and development sets. 

\subsection{Multiple-Choice Datasets}
\textbf{TruthfulQA}~\cite{truthfulqa}: This is a benchmark aimed at evaluating the veracity of language model-generated answers. We obtain it from HuggingFace: \url{https://huggingface.co/datasets/truthful_qa}. 

\textbf{ARC}~\cite{arc_challenge}: The dataset is obtained from the Allen Institute for AI's official website at \url{https://allenai.org/data/arc}. We merge the easy and challenging version of ARC dataset and mark the type in the metadata of each item.

\textbf{HellaSwag}~\cite{hellaswag}: HellaSwag is obtained from the HuggingFace repository \url{https://huggingface.co/datasets/Rowan/hellaswag}. The content from the original data's \texttt{ctx\_a} field is used as the question, while the label field's content is designated as the \texttt{golden\_answers}. Other field information is preserved in the metadata key.

\textbf{MMLU}~\cite{mmlu,mmlu_ethics}: We obtain this dataset in HuggingFace repository \url{https://huggingface.co/datasets/lighteval/mmlu}. In comparison to the standard MMLU dataset, this version includes an auxiliary training set. Given the numerous subsets, each covering various categories of questions, we have merged all subsets into a single dataset. The original MMLU dataset includes a validation set with five samples intended for the 5-shot evaluation scenario. We have separated this set into a distinct file and renamed the original validation set as ``dev''.

\textbf{OpenBookQA}~\cite{OpenBookQA2018}:
The dataset is released by the Allen Institute for AI, which can be downloaded from the URL: \url{https://huggingface.co/datasets/allenai/openbookqa}.  The \texttt{question\_stem} field from the original data is used as the question, while the text field from the choices in the original data is designated as the \texttt{golden\_answers}.

\subsection{Other Datasets}
\textbf{Fever}~\cite{fever}, \textbf{WOW}~\cite{dinan2018wizard}, \textbf{AIDA CoNll-YAGO}~\cite{AIDA_CONLL}, \textbf{WNED}~\cite{wned}, \textbf{T-REx}~\cite{trex}, \textbf{Zero-shot RE}~\cite{zeroshotre}: These datasets are collected from the KILT benchmark~\cite{kilt_2021}. We process the format of the original data file to maintain consistency. Due to the lack of correct answers in the test set, we only retain the training and development sets. For Fever, We put ``SUPPORT'' or ``REFUTES'' into the list as golden answers.

\textbf{WikiAsp}~\cite{wikiasp}: We collect this dataset from the official URL: \url{https://github.com/neulab/wikiasp}. Since the original task is summarization, we adapt it following the approach of FLARE~\cite{jiang2023active} to generate summaries based on the corresponding query and aspect, making it more suitable for the RAG scenario. Specifically, we retrieve the most relevant passages from the Wikipedia corpus using the answer from each data point and use the title of the retrieved passage as the query. Our retriever is \texttt{E5-base-v2}, and the Wikipedia corpus is the DPR version of the 2018 December Wiki. Each data point's aspect is included in the metadata.

\subsection{License}
According to the settings of the original authors for each dataset, the license for each dataset is different, including CC BY-NC-SA 4.0,\footnote{\url{https://creativecommons.org/licenses/by-nc-sa/4.0/}} CC BY-NC-SA 3.0,\footnote{\url{https://creativecommons.org/licenses/by-nc-sa/3.0/}} CC BY 4.0,\footnote{\url{https://creativecommons.org/licenses/by/4.0/}} Apache-2.0,\footnote{\url{https://apache.org/licenses/LICENSE-2.0}} MIT,\footnote{\url{https://choosealicense.com/licenses/mit/}} BSD.\footnote{\url{https://www.linfo.org/bsdlicense.html}} All datasets allow for scientific use and redistribution.

\section{Limitations}
\label{sec:limitations}
Due to limitations in development time and manpower, our toolkit currently has some limitations, which we plan to gradually improve in the future: 

(1) Although we strive to encompass many representative RAG methods, due to time and cost considerations, we have not included all existing RAG works. We have surveyed the RAG works published before 2024 and plan to continue implementing some representative works. However, due to the large quantity, this may require support from the open source community.

(2) Our toolkit lacks support for training RAG-related components. We consider training during the initial design, but given the diversity of training methods and the presence of many repositories specifically dedicated to the training of retrievers and generators, we do not include this part currently. In the future, we plan to add some supplementary scripts to support researchers' training needs.

\end{document}